\documentclass[final,5p,times,twocolumn]{elsarticle}
\usepackage{graphicx} 
\usepackage{balance}
\journal{Pattern Recognition, Vol. 28, No. 12, pp. 1855--1870, 1995}
\begin{document}

\begin{frontmatter}
\title{Polyhedral Object Recognition by Indexing
}
\author{Radu Horaud and Humberto Sossa}
\address{INRIA Grenoble Rh\^{o}ne-Alpes, 38330 Montbonnot Saint-Martin, France}
\begin{abstract}
In computer vision,
the indexing problem is the problem of recognizing a few objects in a
large database of objects while avoiding the help of the classical
image-feature-to-object-feature matching paradigm. In this paper we address
the problem of recognizing 3-D polyhedral objects from 2-D images by indexing.
Both the objects to be recognized and the images are represented by weighted
graphs. The indexing problem is therefore the problem of determining whether a
graph extracted from the image is present or absent in a database of model
graphs. We introduce a novel method for performing this graph indexing
process which is
based both on polynomial characterization of binary and weighted graphs and on
hashing. We describe in detail this polynomial characterization and then we
show how it can be used in the context of polyhedral object recognition. Next
we describe a practical recognition-by-indexing system that includes the
organization of the database, the representation of polyhedral objects in
terms of 2-D characteristic views, the representation of this views in terms
of weighted graphs, and the associated image processing. Finally, some
experimental results allow the evaluation of the system performance.
\end{abstract}


\end{frontmatter}
\section{Introduction}
The problem of object recognition in computer vision is the problem of
matching object features with image features. Nevertheless, since an object
has many features associated with it and since an 
image contains features that do
not necessarily belong to that object, the matching process is a complex
one because of the large size of the set of image-feature-to-object-feature
assignments. Therefore, in the past, the rationale has been to use constraints
-- such as the rigidity constraint -- in order to maintain the number of
assignments as reduced as possible: Therefore, any search process, including
exhaustive search, is likely to be fast enough because it has to visit a few
thousands of nodes rather than millions.

Various implementations of matching-based recognition 
of rigid objects take the form of
either a search-graph or a search-tree. Examples of search graphs are
maximal-clique finding algorithms introduced in computer vision by Ambler \&
al. \cite{Ambler73}, popularized by Ballard \& Brown \cite{BallardBrown82},
and applied to 2-D object recognition by Bolles \& Cain \cite{BollesCain82}.
Subsequently, the advantage of using trees rather than graphs was stressed by
a large number of authors such as Bolles \& Horaud \cite{BollesHoraud86},
Faugeras \& Hebert \cite{FaugerasHebert86}, Ayache \& Faugeras 
\cite{AyacheFaugeras86}, Grimson \& Lozano-Perez 
\cite{GrimsonLozanoPerez87}, Goad \cite{Goad86}, Flynn \& Jain
\cite{FlynnJain91b}, and many others.

However, most of the object recognition methods just mentioned restrict the
recognition to object whose exact geometry is known in advance. A more general
approach consists of representing both the image and the object by graphs and
of casting the recognition problem into the graph matching problem. Graphs are
a convenient way of representing features and relationships between these
features. Various graph representations have been used by Kim \& Kak
\cite{KimKak91}, Flynn \& Jain \cite{FlynnJain91}, Dickinson \& al.
\cite{DickinsonPentlandRosenfeld92}, and Bergevin \& Levine
\cite{BergevinLevine93}. However, graph matching is a difficult problem in
itself. Whenever the two graphs to be matched have the same number of nodes,
graph matching is equivalent to searching for graph isomorphism and polynomial
time solutions exist in this case, \cite{Umeyama88}, \cite{Hanajik93},
\cite{AlmohamadDuffuaa93}. It is however rarely the case that the image graph
have the same size as the object graph: The problem is therefore equivalent to
maximum subgraph matching -- find the largest isomorphic subgraphs of the two
graphs. So far, solutions proposed for solving the maximum subgraph matching
problem involve some form of combinatorial optimization
\cite{HeraultHoraudVeillonNiez90}, \cite{TrespGindi90}.

If many objects rather than a single one (as it has often been the case)
are present in a database of objects to be recognized, then the matching-based
recognition becomes intractable because the complexity grows substantially
with the number of features. An indexing process is crucial whenever
the recognition process has to select among many possible
objects. {\em Recognition by indexing} is
therefore the process that attempts to rapidly extract from a large list of
objects, those few objects that fit a group of image features while avoiding
to establish image-feature-to-object-feature assignments.

Nevatia \& Binford \cite{NevatiaBinford77} were among the first to describe
indexing as a part of an object recognition system. Ettinger \cite{Ettinger88}
described a hierarchically organized object library well suited for indexing.
Each object is decomposed into a list of sub-parts.
The rationale is that many objects share a common set of
sub-parts and what distinguishes one object from another
is sub-part relationships -- the overall list of sub-parts grows
{\it sub-linearly} with the number of objects in the library. This idea is
applied to flat objects that are described by their outlines.

The idea of using hashing in conjunction with object recognition was
introduced
by Kalvin \& al. \cite{Kalvin86}.
Outlines of flat objects are described in
terms of {\it footprints}. The best way to think of a footprint is of an
intrinsic curve such as curvature as a function of curvilinear
abscissa. The footprint of an object is further decomposed into intervals.
Each such interval is described by a set of numbers (the sine and cosine
Fourier coefficients, for example) and these numbers are hashed in
hash-tables. The indexing itself takes the form of a vote: Each footprint
interval detected in the image votes for those objects in the database
containing this footprint interval. Finally the object that received the
highest vote score is the recognized object. A variation of this method using
local frames and {\it geometric} hashing was proposed by Lamdan \& Wolfson
\cite{LamdanWolfson88} for solving the matching problem, not the indexing
problem.

Following the same idea of hashing, Stein \& Medioni \cite{SteinMedioni92}
were able to recognize 3-D objects from 3-D data using super-segments and
surface-patches as features. Their {\it structural} hashing
technique retrieves object hypotheses from the database using hash-table
indexing. A similar approach was proposed by
Breuel \cite{Breuel89}.

The approach advocated in this paper capitalizes onto the representation of
3-D objects in terms of 2-D characteristic views and of characterizing such a
view with a number of weighted graphs. An identical graph representation is
extracted from the image of an unknown 3-D scene. A polynomial
characterization of graphs allows us to organize the ``model graphs" into hash
tables. Therefore, recognition consists of computing similar polynomial
characterizations  for the ``image graphs" and of indexing in the pre-stored
hash tables. Finally, a voting process ranks a number of candidate 
characteristic views as
potential recognized objects.
\subsection{Paper organization}
The remainder of this paper is organized as follows.
Section~\ref{section:polynomial-characterization} introduces the polynomial
characterization of a binary graph that will be used, namely the second
immanantal polynomial of the Laplacian matrix of a graph. Then we briefly
describe an extension of this representation to weighted graphs. Section~\ref{section:graph-indexing} describes the graph indexing method that uses this
polynomial characterization of graphs. It describes as well an object
representation in terms of weighted graphs, the  organization of the database
of objects to be recognized, and the indexing method itself which is based on
hashing. Section~\ref{section:characteristic-views} describes a representation
of 3-D polyhedral objects in terms of 2-D views and a representation of these
views in terms of weighted graphs. Section~\ref{section:image-processing}
describes how to extract these weighted graphs from images and how to remove
irrelevant image data. Section~\ref{section:experiments} describes a
recognition experiment carried out with a set of 9 images of the same scene in
the presence of a database of 6 objects. Finally,
section~\ref{section:discussion} draws some conclusions and gives some
directions for future work.
\section{Polynomial characterization of a graph}
\label{section:polynomial-characterization}

The method that we propose in this paper in order to achieve indexing uses
graphs to represent both images and objects. Let us suppose that one is able
to extract a number of graphs from an image and let $G_1$ be such an image
graph that has the same number of nodes as a graph $G_2$ extracted from an
object. Such a graph (an image or an object graph)
is defined by a set of vertices $V$ and a set of edges
$E$. The two graphs $G_1=(V_1,E_1)$ and $G_2=(V_2,E_2)$ 
are said to be isomorphic if there is a bijection 
$\phi:V_1 \longrightarrow V_2$ such that:
\[ (v_1,v_2) \in E_1 \;\; \mbox{if and only if} \;\; (\phi(v_1),\phi(v_2)) \in E_2
\]

If $A_1$ and $A_2$ are the adjacency matrices of the two graphs, one can easily
see that $G_1$ is isomorphic to $G_2$ if and only if there exists a
permutation matrix $P$ satisfying:
\begin{equation}
\label{eq:similarity-transformation}
A_2 = P A_1 P^{-1}
\end{equation}

Hence, there are two ways to decide whether two graphs are isomorphic:
\begin{enumerate}
\item Find the permutation matrix $P$ that satisfies the equation above. 
\item Find an algebraic characterization of the adjacency matrix of a graph
that is {\em invariant} under a similarity transformation of the adjacency
matrix. Such a characterization has been proved to be useful for graph
classification. 
\end{enumerate}
One obvious characterization that is invariant under similarity is the
characteristic polynomial associated with the adjacency matrix
\cite{CvetkovicDoobSachs80}, \cite{Turner68}. Indeed we have:
\begin{eqnarray*}
det(xI -  P A P^{-1}) & = & det(PxIP^{-1} -  P A P^{-1}) \\
		      & = & det(P(xI - A)P^{-1}) \\
		      & = & det(xI - A)
\end{eqnarray*}
Therefore, the similarity of adjacency matrices is a necessary condition for
isomorphism. Unfortunately it is far from being a  sufficient condition.
However, an important idea stems out from this example of graph characterization
-- one may seek to characterize a graph, up to an isomorphism, by the
coefficients of a polynomial associated with that graph. More formally, we
seek a polynomial associated with a graph, say $p(G)$ such that:
\begin{equation}
        \left\{
        \begin{array}{l}
\mbox{if $G_{1} = G_{2}$ then $p(G_{1}) = p(G_{2})$} \\
\mbox{and}\\
\mbox{if $p(G_{1}) = p(G_{2})$ then $G_{1} = G_{2}$}
        \end{array}
        \right.
\label{eq:conds}
\end{equation}

Two graphs are said to be equal if they have the same number of nodes and if
they are isomorphic. Two polynomials are equal if they have the same degree
and if their coefficients are equal. 
If a polynomial satisfying the above condition exists, it follows that
the problem of comparing two graphs of the same size 
is equivalent to the
problem of comparing the coefficients of their associated polynomials.
Notice however that graph characterization with a polynomial allows one to
state whether two graphs are isomorphic or not but it doesn't provide the
node-to-node isomorphic mapping between the graphs. Graph characterization is
therefore exactly what one needs for model indexing, i.e., rapidly state
whether some sensed data {\it equal} some object data. The search of an
isomorphic node-to-node mapping is the task of matching and not the
task of indexing.

Polynomials that characterize a graph unambiguously up to an isomorphism have
been thoroughly studied in the linear algebra literature \cite{Turner68},
\cite{MerrisRebmanWatkings81}. Among these
polynomials, the {\em second immanantal polynomial} -- or the
$d_{2}$-polynomial -- is a good candidate \cite{Merris86}. 

One may associate the second immanantal polynomial with the adjacency matrix
of a graph. There are however some reasons to prefer the Laplacian matrix
(defined below) to the computationally simpler adjacency matrix. The Laplacian
matrix is positive semidefinite symmetric of rank $n-1$ (if $G$ is a connected
graph). Generally speaking, second immanantal polynomials match up well with
positive semidefinite matrices. The greater complexity of Laplacian matrices,
when compared with adjacent matrices, suggests there may be fewer {\em
algebraic accidents}  \cite{Merris86}. If the time to compute the determinant
of a $n\times n$ matrix is $n^3$, the time to compute the coefficients of the
second immanantal polynomial is $n^4$.

The elements of the 
Laplacian matrix of a {\em binary graph}, $L(G)$, are defined as follows:
\begin{equation}
        l_{ij} = \left\{
                        \begin{array}{lll}
                        d_{i} & \mbox{if $i = j$} \\
                        -1 & \mbox{if there is an edge between nodes $i$ and
$j$}
 \\
                        0 & \mbox{otherwise}\\
                        \end{array}
                \right.
\label{eq:matlap}
\end{equation}
where $d_{i}$ is the number of graph edges meeting at node $i$ 
and is called the degree of the node $i$. The interest reader may find 
in \cite{Constantine90} a complete description of the properties of the 
Laplacian matrix of a binary graph.

The second
immanantal polynomial associated with a $n\times n$ Laplacian matrix of a
graph, 
$L(G)$, can be written in generic form as:
\begin{eqnarray}
d_{2}(xI - L(G)) &=& c_{0}(L(G)) x^{n} - c_{1}(L(G)) x^{n-1} + ... \nonumber \\
&+& (-1)^{n} c_{n}(L(G))
\end{eqnarray}
The coefficients $c_{o},...,c_{n}$ of this polynomial are
integers and they can be computed using
the following formulae which are detailed by Merris
\cite{Merris86} ($n$ is the number of nodes of the
graph and $m$ is the number of edges of the graph):
\begin{equation}
        \left\{
        \begin{array}{lll}
c_{0}(L(G)) & = & n-1 \\
c_{1}(L(G)) & = & 2m(n-1) \\
         & \vdots & \\
c_{k}(L(G)) & = & \sum_{X \in
Q_{k,n}}^{}\big(\sum_{i=1}^{n}~l_{ii}~det~(L(G)\{X\}
(i))  \\
&&- det~(L(G)\{X\})\big)
        \end{array}
        \right.
\label{eq:coefsd2}
\end{equation}
In these formulae $l_{ii}$ denotes a diagonal term of $L(G)$ and
$Q_{k,n}$ denotes the set of all the $C_{n}^{k}$ strictly increasing
sequences of size $k$ ($2\leq k\leq n$) obtained from the set $\{1,2,...,n\}$.
For any $n\times~n$ matrix $M$ and for $X\in~Q_{k,n}$ let
$M[X]$ be the $k\times~k$ principal sub-matrix of $M$ corresponding to $X$.
$M\{X\}$ is the $n \times n$ matrix:
\begin{eqnarray}
M\{X\} = \left(\
                \begin{array}{cc}
                M[X] & 0_{k} \\
                0_{k} & I_{n-k}
                \end{array}
        \right)
\label{eq:prinsubm}
\end{eqnarray}
where $I_{n-k}$ is the identity matrix of size $n-k$ and $0_{k}$ is the null
matrix of size $k$. $M\{X\}(i)$ is the matrix obtained from $M\{X\}$ by
removing the $i$-th row and the $i$-th column.

An important property of the second immanantal polynomial associated with a
graph is that it is preserved under similarity permutation
\cite{MerrisRebmanWatkings81}:
\[
d_2(xI - L(G)) = d_2(xI - PL(G)P^{-1})
\]
Therefore, a necessary condition for two graphs to be isomorphic is that they
have the same second immanantal polynomial. However, it is not a sufficient
condition. In practice, however, there have been found very few examples of
non-isomorphic graphs that have the same second immanantal polynomial 
\cite{CvetkovicDoobSachs80}.

\subsection{An example}

\begin{figure}[h]
%\centerline{\psfig{figure=../two-graphs.idraw}}
\centerline{\includegraphics[width=0.49\textwidth]{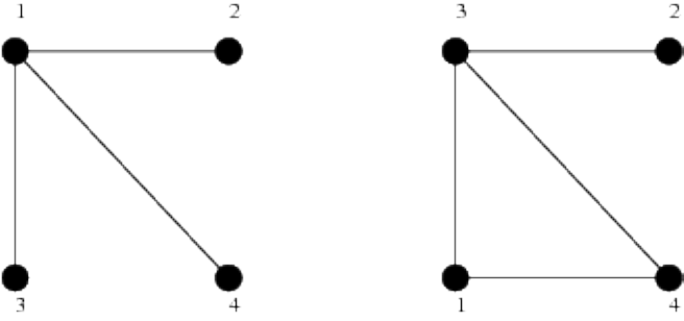}}
\caption{These two graphs differ by one edge but their associated second
immanantal polynomials are quite different.}
\label{fig:two-graphs}
\end{figure}

In order to illustrate the above formalism let us consider two simple binary
graphs and let us outline the computation of their associated second immanantal
polynomials. An example of two 4-node binary graphs are shown on
Figure~\ref{fig:two-graphs}. The Laplacian matrices are given by
equation~(\ref{eq:matlap}) and they are easy to compute:
\begin{eqnarray*}
L(G)_1 & = & \left( \begin{array}{rrrr}
		    3 & -1 & -1 & -1 \\
		   -1 &  1 &  0 &  0 \\
		   -1 &  0 &  1 &  0 \\
		   -1 &  0 &  0 &  1
		    \end{array}
	     \right)			\\
L(G)_2 & = & \left( \begin{array}{rrrr}
                    2 &  0 & -1 & -1 \\
                    0 &  1 & -1 &  0 \\
                   -1 & -1 &  3 & -1 \\
                   -1 &  0 & -1 &  2
                    \end{array}
             \right)
\end{eqnarray*}

For $n=4$ the sets of all the $C_{4}^{k}$ strictly increasing
sequences of size $k$ ($2\leq k\leq 4$) are:
\begin{eqnarray*}
Q_{2,4} & = & \{(1,2),(1,3),(1,4),(2,3),(2,4),(3,4)\} \\
Q_{3,4} & = & \{(1,2,3),(1,2,4),(1,3,4),(2,3,4)\} \\
Q_{4,4} & = & \{(1,2,3,4)\}
\end{eqnarray*}

It is straightforward to compute the matrices $L(G)\{X\}$ and the matrices
$L(G)\{X\}(i)$, for $X\in Q_{k,4}$. For example, $L(G)_1\{(1,2)\}$ is a
4$\times$4 matrix obtained by appending the first two rows and first two
columns of $L(G)_1$ with $I_2$ and $0_2$ as follows:
\[
L(G)_1\{(1,2)\} = \left( \begin{array}{rrrr}
			3 & -1 & 0 & 0 \\
		       -1 &  1 & 0 & 0 \\
			0 &  0 & 1 & 0 \\
			0 &  0 & 0 & 1
			\end{array}
		 \right)
\]
$L(G)_1\{(1,2)\}(2)$ is a 3$\times$3 matrix obtained from $L(G)_1\{(1,2)\}$ by
removing its 2nd row and 2nd column:
\[
L(G)_1\{(1,2)\}(2) = \left( \begin{array}{rrr}
			3 & 0 & 0 \\
			0 & 1 & 0 \\
			0 & 0 & 1
                        \end{array}
                 \right)
\]
After some straightforward computation we obtain the following coefficients
for the associated polynomials:
\begin{eqnarray*}
d_{2}(xI - L(G)_1) & = & 3x^4 - 18x^3 + 33x^2 - 24x + 6 \\
d_{2}(xI - L(G)_2) & = & 3x^4 - 24x^3 + 105x^2 - 68x + 24
\end{eqnarray*}

One may also compute the characteristic polynomials associated with the
Laplacian matrices, i.e.:
\begin{eqnarray*}
det(xI - L(G)_1) & = & x^4 - 6x^3 + 9x^2 - 4x \\
det(xI - L(G)_2) & = & x^4 - 8x^3 + 19x^2 - 12x
\end{eqnarray*}

From this example it is obvious that the second immanantal polynomial is a
richer graph description than just the characteristic polynomial.
\subsection{Weighted graphs}
In general, binary graphs are not sufficient for describing the structure of
either images or objects. Weighted graphs are graphs which have a weight
$w_{ij}$ associated with the edge linking nodes $i$ and $j$. The definition of
the Laplacian matrix may easily be extended to weighted graphs, as follows:
\begin{equation}
        l^{w}_{ij} = \left\{
        \begin{array}{lll}
        D_{i} & \mbox{if $i = j$} \\
        -w_{ij} & \mbox{if there is a weighted edge between $i$ \& $j$} \\
        0 & \mbox{if there is no edge between nodes $i$ \& $j$}
                        \end{array}
                \right.
\label{eq:newmatlap}
\end{equation}
with $D_{i}$ being equal to the sum of the weights of the edges meeting at
the node $i$:
\begin{equation}
D_{i} = \sum_{j=1}^{n}~w_{ij}
\end{equation}

This matrix has the same properties as the Laplacian matrix associated
with a binary graph -- it is symmetric semidefinite positive and of rank
$n-1$ which makes it
suitable for computing the $d_{2}$-polynomial.

\section{Graph indexing}
\label{section:graph-indexing}

The graph characterization in terms of the coefficients of the second
immanantal polynomial allows one to assert whether two graphs with the same
number of nodes ($n$) are ``equal''.
The difference between two graphs $G_{1}$ and $G_{2}$ is given by the
formula:
\begin{equation}
\mbox{Diff}(G_{1},G_{2}) = \sum_{k=1}^{n}~(c_{k}^{1} - c_{k}^{2})^{2}
\end{equation}
where $c_{k}^{i}$ is the $k$-th coefficient of the second immanantal
polynomial associated with graph $i$. Since we assume that the above equation
is valid only for graphs with the same number of nodes, $c_0$ has been skipped
out from the summation.

In the case of indexing we are faced with the problem of comparing an
image graph with many database graphs and of deciding which are the few
graphs in the database that are equal to the image graph.
In that case the graph difference mentioned above is not efficient.
One way to implement indexing efficiently is to use {\em hashing}
\cite{Sedgewick88}.
Hashing can be briefly described
as follows. Each database object has a numerical {\em key} associated
with it. Then a {\em hash function} maps this key onto the address of an
array of a manageable size. The address thus computed for an object is also
called the {\em hash-code} of that object. In practice,
hashing is composed of an off-line step
(database construction) and a runtime step (indexing):
\begin{itemize}
\item {\it Database construction} consists of computing a hash-code 
for each object to be stored in the database. Several objects may well have the same address (hash-code). Therefore a list
of objects will be associated with each address.
The database takes therefore the form of an array (or a hash-table), 
a list of objects being stored at each array-element address.
\item {\it Indexing} consists of computing
the address (hash-code) of an unknown object in
order to determine whether this object is in the hash-table or not.
\end{itemize}

Since a graph may be described by the integer 
coefficients of a polynomial, these coefficients may 
well be viewed as the hash-codes of
the graph. Hence, a graph with $n$ nodes can be mapped onto $n$ hash tables.
For reasons that will be made clear below, the size of the graphs we deal
with varies between 5 nodes and 10 nodes. Within this size range the second
immanantal polynomial uniquely characterizes binary and weighted graphs. It
follows that graph indexing will become an efficient technique because the
hashing will have very few collisions associated with it.

Polynomial characterization of graphs combined with graph indexing will
eventually allow us to perform object recognition by indexing. However, two
important issues need be raised before we describe a practical object
recognition system: object representation and database organization.

\subsection{Object representation}

Object representation has been thoroughly studied in Computer
Vision and a recent paper by Flynn \& Jain \cite{FlynnJain91} provides a good
state of the art. In general there are two possible representation classes:
Object frame centered and viewer frame centered representations. 
Within our approach we
use a representation that is not tight to a specific coordinate
frame. An object is
mainly described by a list of characteristic views. In the representation that
we use the definition of a characteristic view (CV) should be understood in a
broad sense. It is a network of object features and feature relationships that
are simultaneously visible from some viewpoint. Such a representation is by no
means limited to the aspect graph representation of an object. The features in
the network may well be either 2-D or 3-D, object-centered or viewer-centered.
The important characteristic here is not as much the dimensionality of the
features or the coordinate
frame to which they relate, but instead, the {\it intrinsic}
properties of the feature network.
As we already mentioned, such a feature network can be conveniently
represented by a weighted graph.

However, the data associated with some view of an object rarely encodes a
whole characteristic view associated with that object. The data are corrupted
by noise, occlusions, self occlusions, and accidental alignments. Therefore
it will not be very useful to directly store in the database the graph
associated with a characteristic view. Instead, each characteristic view is
further decomposed in a number of, possibly overlapping, ``smaller" views or
subviews,
where each such subview is in fact associated with a subgraph of the graph
describing the characteristic view. There are several reasons in support of
the decomposition of a characteristic view into a number of subviews:
\begin{itemize}
\item Following the results of
section~\ref{section:polynomial-characterization}, one can compare only graphs
with the same number of nodes. Since a graph extracted from the data has
rarely the same number of nodes as the graphs associated with the
characteristic views of the objects to be recognized, one may still attempt to
compare an unknown-object-view with a characteristic-view by comparing
subgraphs associated with subviews of theses views.
\item The cost of the computation of the coefficients of the second immanantal
polynomial is proportional to 
$n^4$, where $n$ is the number of nodes of the graph. Since ``raw"
characteristic views may have a large number of features associated with them, 
it may not be efficient to compute polynomial characterizations for very
large graphs. 
\item Consider a data graph that is composed of a large network of features.
It is very unlikely that such a large data graph belongs to a unique object.
Recognition based on such large graphs will fail because these graphs are not
present in the database. 
\end{itemize}

\begin{figure}[t!]
%\centerline{\psfig{figure=../database.idraw}}
\centerline{\includegraphics[width=0.49\textwidth]{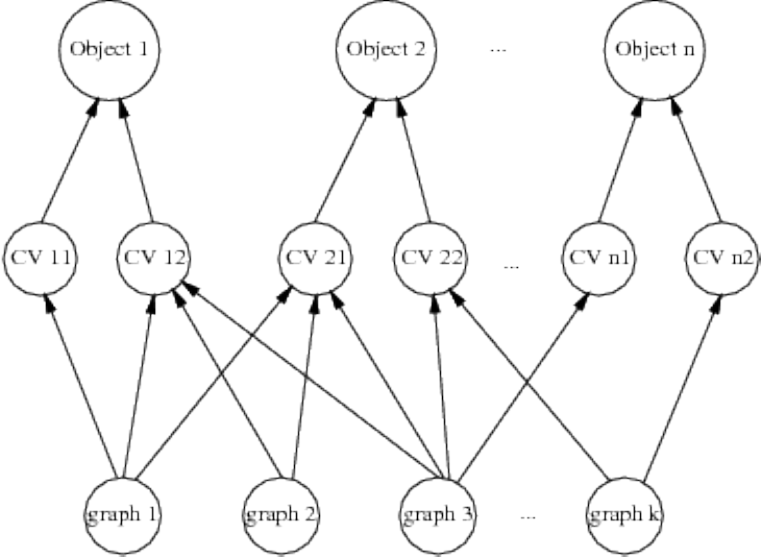}}
\caption{The database has a three layer structure: graphs, characteristic
views, and objects. An object may well have more than 2 characteristic views
associated with it.}
\label{fig:database}
\end{figure}

It is therefore clear that object recognition by indexing must adopt a
representation such that each characteristic view is decomposed in a number of
subviews or subgraphs. 
A compromise must be made concerning the size of these subgraphs:
Very small subgraphs are too ambiguous because they do not capture information
that is object specific while large subgraphs are difficult to extract from
the data. At the limit, a single-node graph belongs to all the objects in
the database and indexing is useless in this case. At the other extreme, a
very large graph encapsulates more than one object and the indexing process
will fail to find this graph in the database.

\subsection{Database organization and the indexing mechanism}

Following the above discussion, the organization of the database follows a
three-layer structure. A first layer contains a list of graphs of various
sizes that are organized in hash tables. The second layer contains
characteristic views. A third layer contains descriptions of the object
themselves. This structure is shown on Figure~\ref{fig:database}.

It is clear that a graph belongs to several characteristic views and hence, it
may belong to several objects. Therefore, an image graph that matches a graph
in the database provides handles to more than one object. The interesting
feature of this three-layer organization is that the graph list grows
sub-linearly with the number of characteristic views.

The indexing mechanism proceeds as follows. Let's suppose that an unknown
object view has to be recognized. First, this unknown view is decomposed into
subviews and a graph is associated with each subview. Polynomial
characterizations are computed for these unknown graphs. Based on these
characterizations and on the hashing technique just described, each unknown
graph is assigned a {\bf unique} graph in the database. As a consequence, a
list of characteristic views may now be associated with each unknown graph in
the image. In other terms, each unknown graph {\em votes} for a number of
characteristic views. This process is repeated for each unknown graph
belonging to the unknown view. The characteristic view that received the
largest number of votes is the model that best matches the unknown view.

Consider, for example the database depicted on Figure~\ref{fig:database} and
suppose that two unknown graphs are assigned {\tt graph1} and {\tt graph2}
respectively. It follows that two characteristic views ({\tt CV12} and
{\tt CV21}) received 2 votes while one characteristic view ({\tt CV11}) received
only one vote.

\begin{figure}[t!]
%\centerline{
%\psfig{figure=../FigffL.idw}}
\centerline{\includegraphics[width=0.49\textwidth]{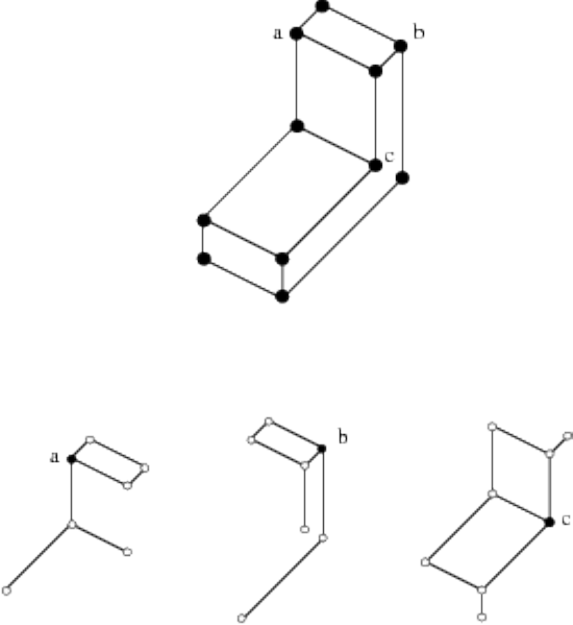}}
\caption{An example of a characteristic view of an object and a few
graphs extracted from this view.}
\label{fig:ff_subgraphs}
\end{figure}

\section{2-D characteristic views of polyhedral objects}
\label{section:characteristic-views}

In order to recognize an object with the method described above, one has to
represent that object in terms of a few characteristic views and to describe
each such characteristic view in terms of a set of weighted graphs. Unlike
solutions that consists of computing characteristic views from CAD object
descriptions, our approach for obtaining these views is to gather as many
images of an object as characteristic views are needed for describing that
object unambiguously. Although in this paper we use an ad-hoc technique, more
formal methods may be found in \cite{Gros92c}, \cite{Gros93d}: 
The authors define a set of characteristic views of a polyhedral object
by partitioning a large set of object views into small sets of
characteristic views.

The task of decomposing a characteristic view into subgraphs is not an easy
one. The most general approach would be to implement a graph decomposition
method. For example, one may attempt to partition a graph into a pre-specified
number of subgraphs such that the number of connections (edges) between the
subgraphs is minimized \cite{HeraultNiez89}. Here we prefer a more pragmatic
solution. For example, one may consider all the nodes of a characteristic view
and form subgraphs around each such node. A subgraph is thus formed by this
node as well as the nodes that are at a distance less or equal than $p$ edges
away from this node. It turns out that this redundant  decomposition of a
characteristic view in small graphs of various sizes is one key to the success
of our recognition method. Indeed, small perturbations in the topology of a
view (due essentially to noise or to segmentation errors) will not affect the
topology of {\em all} the subgraphs extracted from this view.

In the particular case of polyhedral objects, if the degree of a node is, on
an average, equal to 3 and for $p\leq 2$ then the number of nodes of the
associated subgraphs varies between 5 and 10. Figure~\ref{fig:ff_subgraphs}
shows a characteristic view of a simple polyhedral object and some subgraphs
extracted from this view with $p=2$.

\begin{figure}[t!]
%\centerline{
%\psfig{figure=../Figsamestr.idw}}
\centerline{\includegraphics[width=0.49\textwidth]{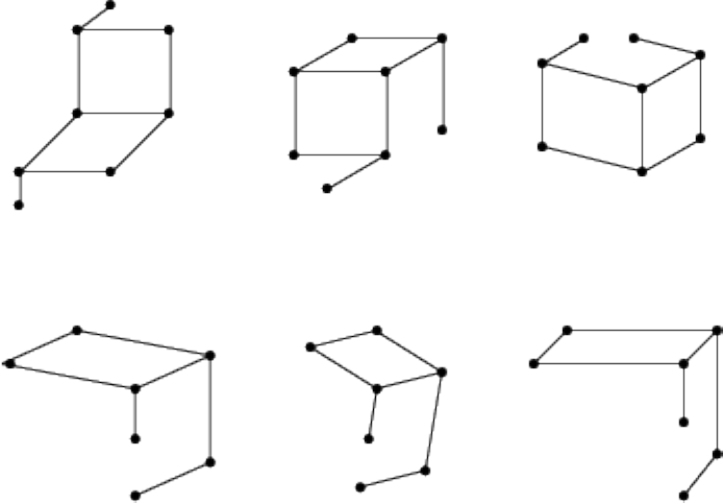}}
\caption{Top -- three graphs having the same topology. Bottom --
three other graphs having the same topology and the same appearance.}
\label{fig:sametopo_samest}
\end{figure}

As it has already been discussed, the topology of a characteristic view is not
sufficient for describing the view unambiguously. For example,
Figure~\ref{fig:sametopo_samest} shows six different binary graphs which are
isomorphic (they have the same topology). Clearly, one would like to
be able to state that the top three graphs are different and the
bottom three ones are identical. In other
words, the top three ones do not look the same, although they have the same
topology. The question of how to describe the 2-D appearance of a polyhedral
object has already been addressed (see for example \cite{Kanade81}) 
but the question
of how to represent such an appearance with a weighted graph has not. 

\begin{figure}[htb]
%\centerline{
%\psfig{figure=../Figredbsff.idw}}
\centerline{\includegraphics[width=0.49\textwidth]{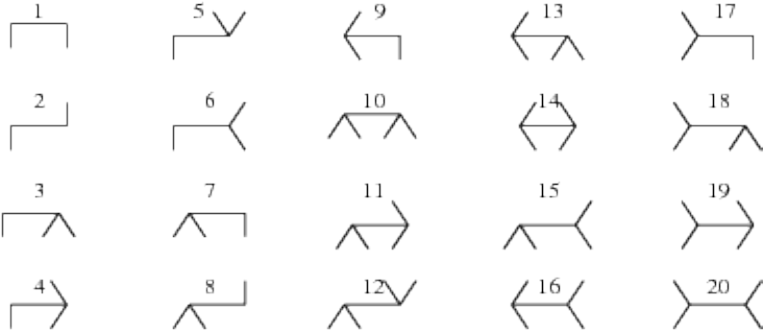}}
\caption{An exhaustive catalogue of the possible appearances of the edges of a
polyhedral object that has, at most, 3 edges meeting at a vertex.}
\label{fig:bsff_ordonees}
\end{figure}

One way to label an edge is to characterize it according to the structure of
the vertices at each endpoint of that edge. If we consider polyhedra that have
at most 3 edges meeting at a vertex, then we obtain a catalogue of possible
edge structures or {\em edge appearances}. It is sufficient to assign labels
to these various appearances and to associate a weight to each label.
Figure~\ref{fig:bsff_ordonees} shows an exhaustive 
catalogue of edge appearances and their weights.

\begin{figure}[htb]
%\centerline{
%\psfig{figure=../edge-struct.idraw,width=12cm,height=8cm}}
\centerline{\includegraphics[width=0.49\textwidth]{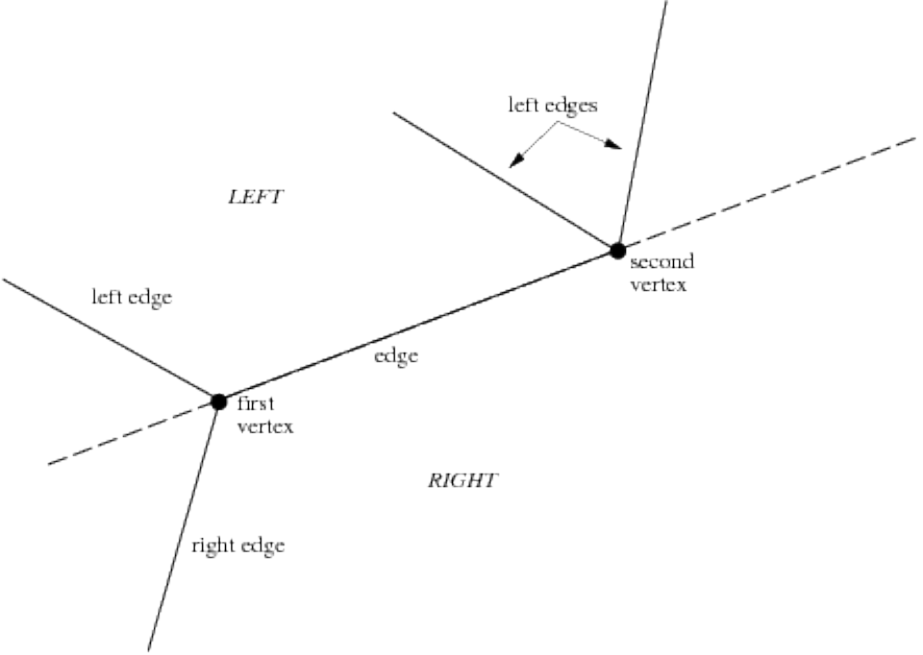}}
\caption{The labelling of an edge depends on the structure of the two vertices
at each endpoint of that edge (see text).}
\label{fig:edge-structure}
\end{figure}

There are three possible vertex structures: a 2-edge vertex or an ``L",
and two 3-edge vertices, an ``Y" and an ``Arrow". Since an edge divides the
plane into two regions -- the left side and the right side -- we obtain the
following list of features that allows the labelling and the
weighting of an edge (see
Figure~\ref{fig:edge-structure} for an example of an edge labelled ``15"):
\begin{itemize}
\item the type of the first vertex ({\tt Arrow});
\item the type of the second vertex ({\tt Arrow});
\item the number of edges associated with the first vertex lying on the
left side of the edge; ({\tt 1})
\item the number of edges associated with the first vertex lying on the right
side of the edge ({\tt 1});
\item the number of edges associated with the second vertex lying on the
left side of the edge ({\tt 2});
\item the number of edges associated with the second vertex lying on the right
side of the edge ({\tt 0}).
\end{itemize}

\section{Image processing}
\label{section:image-processing}

In this section we describe the process by which a number of graphs is
extracted from an image. This graph extraction process has some similarities
with feature grouping since its goal is to provide a few ``key" image features
and reduce the complexity of the object recognition process. Image processing
starts with extracting edges and with approximating these edges with straight
lines. Junctions are next extracted. The junctions and the straight lines form
a network of features, or a graph -- the image graph.

If the scene is composed of just one object, then this image graph
corresponds, up to some noise, to a view of that object. 
Single object scenes are used
for building the database of characteristic views and it has been previously
described.

If the scene is composed of more than one object and if the background is not
uniform, then the image graph has to be further processed in order to be split
into smaller graphs. Each small graph thus obtained is examined in order to
decide whether it should be considered for recognition or not. To summarize,
the process of extracting graphs from an image comprises the following steps:
\begin{enumerate}
\item image graph extraction;
\item image graph splitting, and
\item graph evaluation.
\end{enumerate}

\begin{description}
\item[Step 1.] Image graph extraction has been briefly outlined at the
beginning of this section and is described in detail in
\cite{HoraudVeillonSkordas90}. 
\item[Step 2.] Image graph splitting is
based on a number of heuristics:
\begin{itemize}
\item[2.1] Isolated and ``dangling" edges are thrown out.
\item[2.2] It is assumed that ``T" junctions arise from occlusions (an
object in front of another object, an object in front of some background, or a
self occlusion). Hence, the image graph is cut off at T junctions. Notice that
this process may produce isolated edges which are immediately thrown out.
\item[2.3] Sequences of collinear edges are assumed to arise from the same
physical edge and hence, collinear edges are fused into a unique edge.
\item[2.4] Finally, the image graph is decomposed into connected components.
\end{itemize}

\item[Step 3.] Graph evaluation considers each connected component, one by
one, and evaluates it in order to decide whether it should be further
considered for recognition or thrown out. Let $n$ be the number of nodes of a
graph and let $d_i$ be the degree of node $i$, 
i.e., equation~(\ref{eq:matlap}).
The quantity:
\[	f(G) = \frac{1}{n} \; \sum_{i=1}^{n} d_i	\]
allows one to measure the complexity of a graph. It is straightforward to
notice that for $f(G)=2$ the graph has at most one cycle. Since graphs without
cycles are not really relevant, one may consider only graphs for which:
\[	f(G) \geq 2	\]
The graphs that don't satisfy this constraint are consider irrelevant and
therefore they are thrown out.
\end{description}

\begin{figure}[t!]
\centerline{\includegraphics[width=0.49\textwidth]{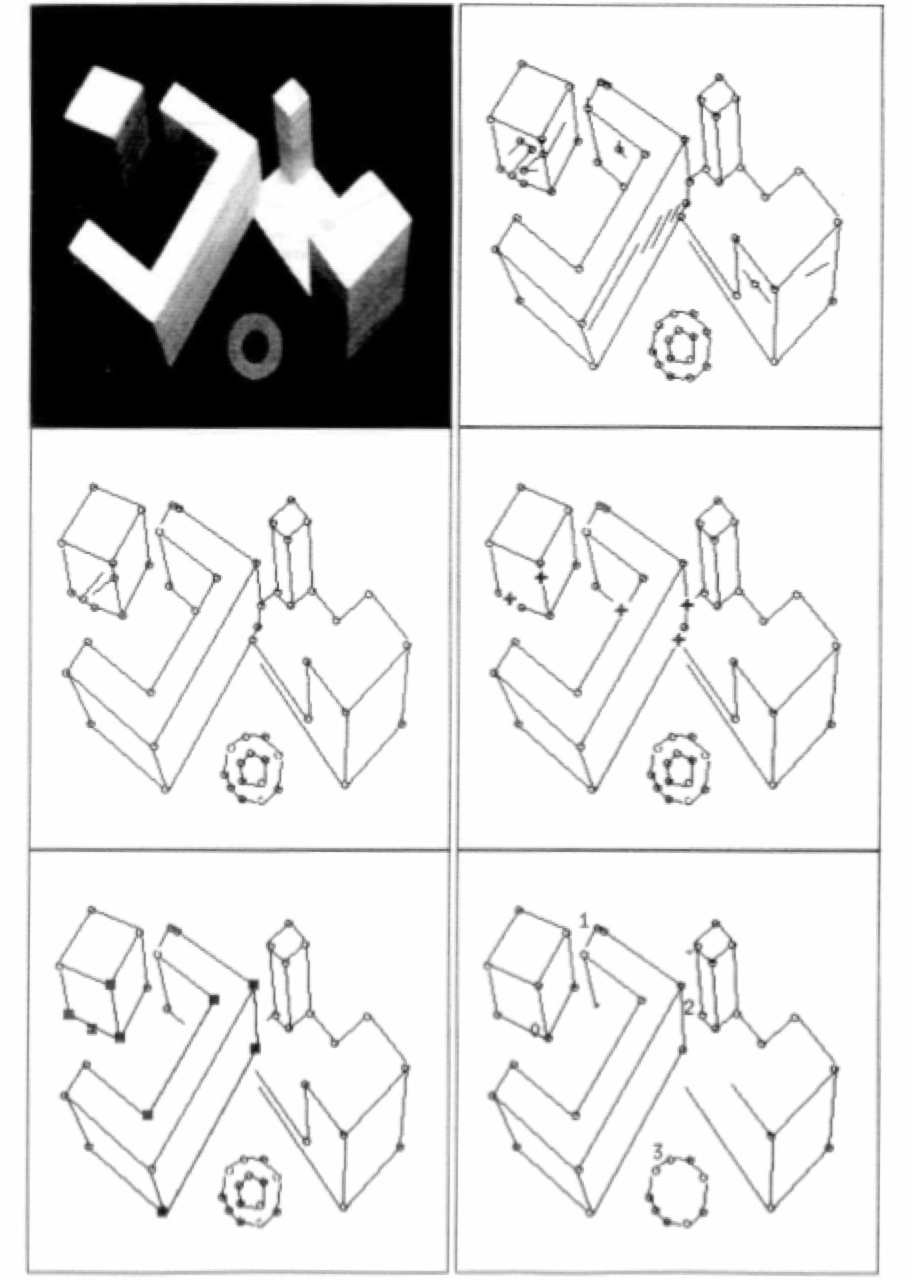}}
\caption{An example of extracting a set of 4 image graphs from an intensity
image (top-left): Lines and junctions are detected (top-right),
isolated and dangling lines are
removed (middle-left), T-junctions are detected and removed (middle-right),
collinear lines are fused into longer lines (bottom-left), the remaining
graphs are evaluated and four of them survive (bottom-right).}
\label{fig:example-processing}
\end{figure}

Let us illustrate with an example the graph extraction process that we just
described. Figure~\ref{fig:example-processing} shows an intensity image
(top-left) and a network of lines and junctions extracted from this image
(top-right) from which isolated and dangling
lines are removed (middle-left). The next
image (middle-right) shows the T-junctions that are removed from the list of
junctions. This T-junction removal 
process produces isolated lines and dangling lines on one hand 
(which are removed) and
collinear lines which are fused into a unique line on the other hand
(bottom-left). We are left now with a number of connected image graphs. Each
such connected component is evaluated according to the 
Step 3 just above. The latter process leaves 4 connected components in the
image (bottom-right). These remaining image graphs will be further 
decomposed into subgraphs in order to be used by the indexing process. The
decomposition of the 4 image graphs into subgraphs is not shown.

\begin{figure}[t!]
\centerline{\includegraphics[width=0.5\textwidth]{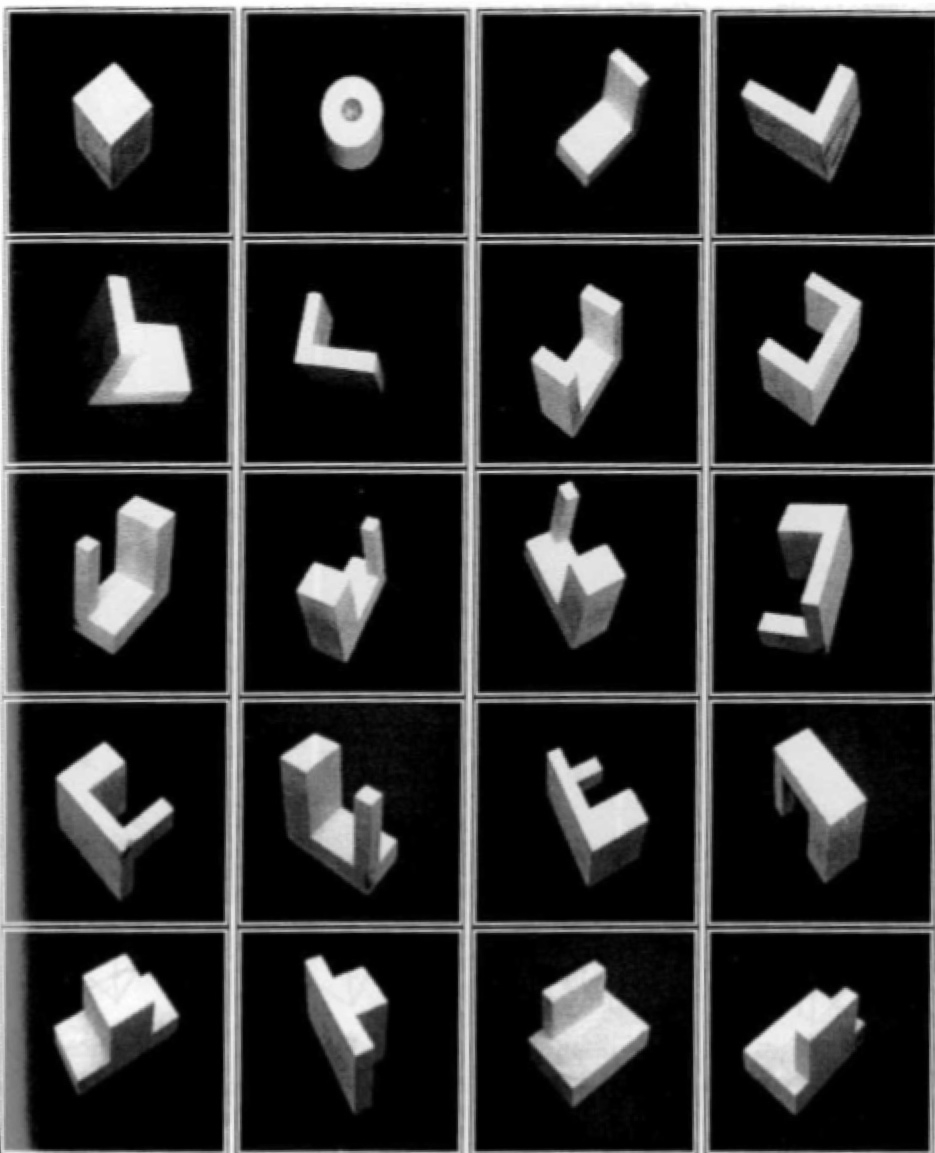}}
\caption{This figure shows the intensity images of 20 characteristic views
associated with 6 objects. These views and objects constitute the database.}
\label{fig:vue-caract}
\end{figure}

\begin{figure}[t!]
\centerline{\includegraphics[width=0.5\textwidth]{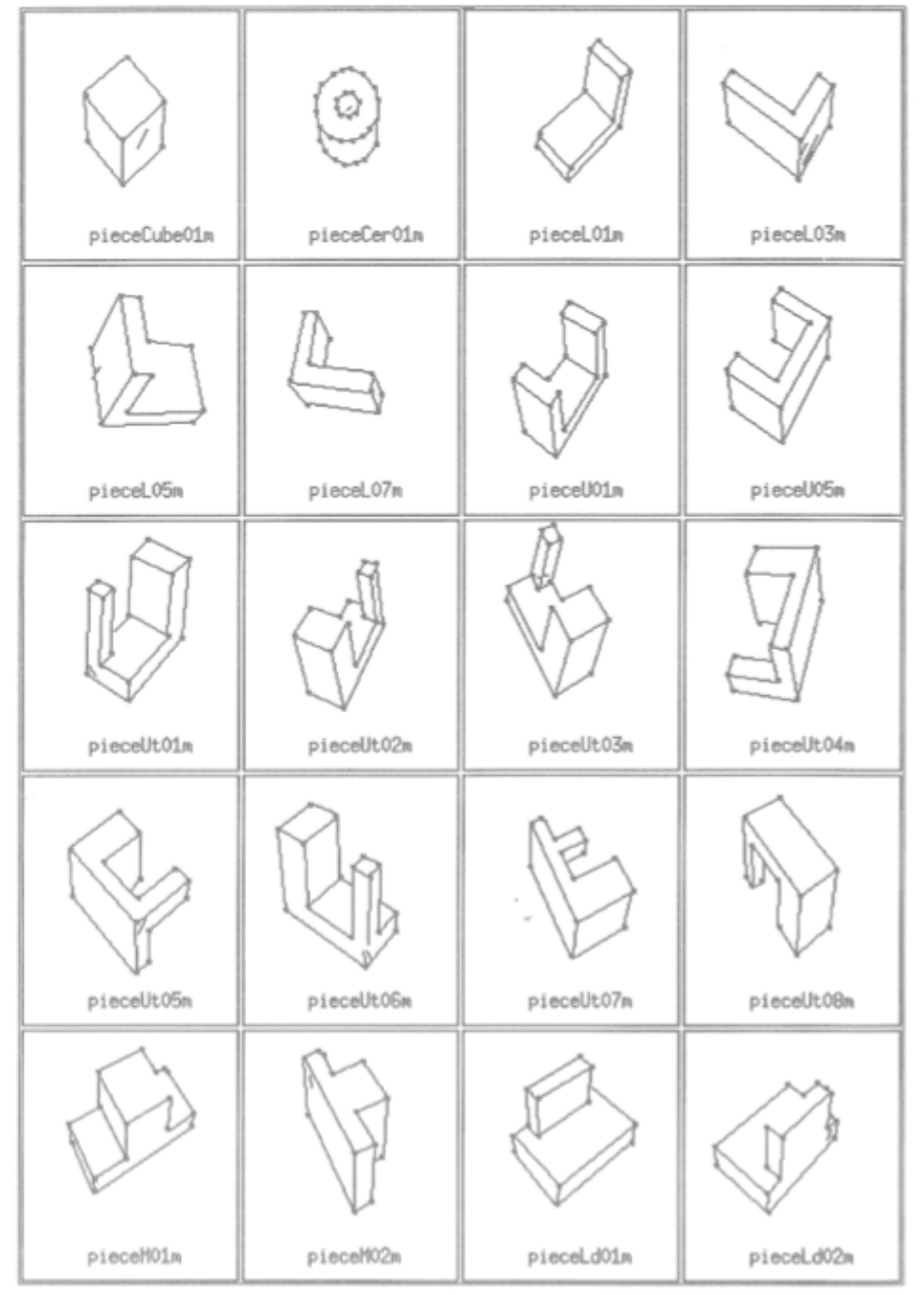}}
\caption{The 20 graphs (lines and junctions) extracted from the previous
images. These 20 graphs are decomposed into subgraphs (the decomposition is
not shown) and stored in the hash-tables associated with the database.}
\label{fig:vue-caract-edges}
\end{figure}

\section{Experiments}
\label{section:experiments}

All the object recognition experiments that we performed used the same
database, namely 20 characteristic views associated with 6 objects, as shown
on Figures~\ref{fig:vue-caract} and \ref{fig:vue-caract-edges}. All the
objects in this database are 3-D polyhedral shapes with one exception. The
database was built by showing each object, view by view, to the camera and by
applying the graph extraction process described in 
section~\ref{section:image-processing}.

We carried out many experiments in which the input image varied from a very
simple one with just one object against an uniform background to more complex
images with many objects against a non-uniform background. 
% All these experiments are reported in \cite{Sossa92}.

\begin{figure}[t!]
\centerline{\includegraphics[width=0.5\textwidth]{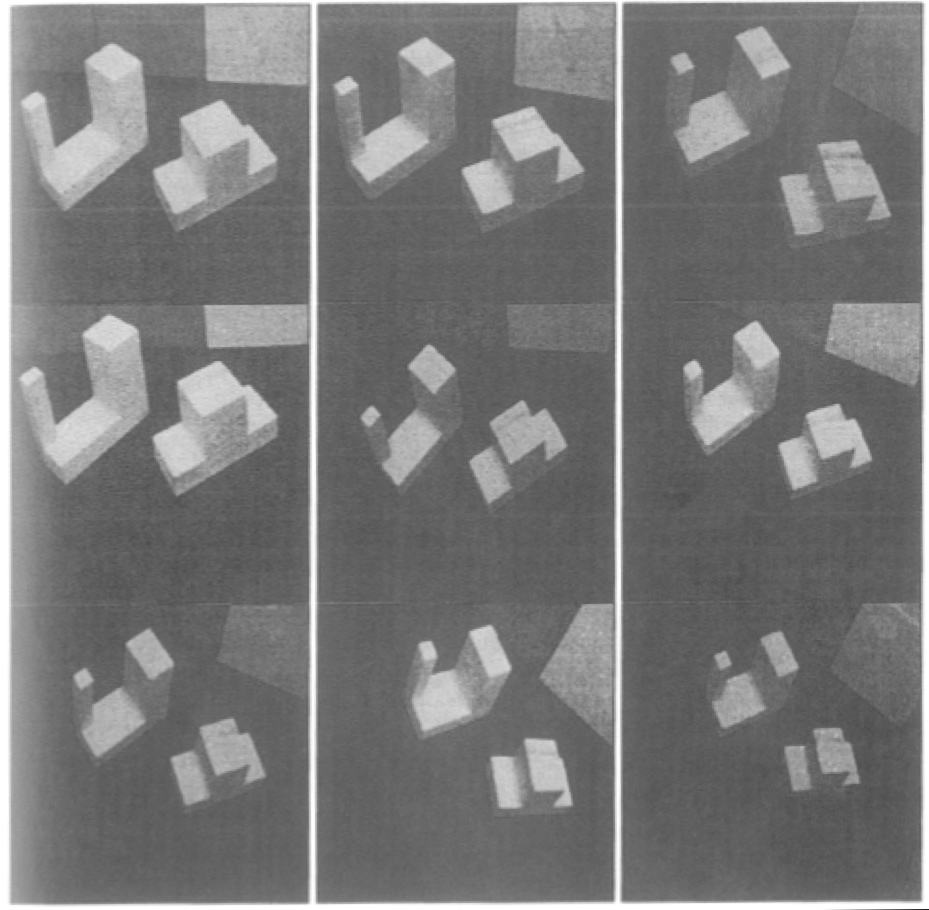}}
\caption{Six images of two objects to be recognized where the camera varies in
position and orientation with respect to the two objects.}
\label{fig:images-exp}
\end{figure}

\begin{figure}[h!]
\centerline{\includegraphics[width=0.5\textwidth]{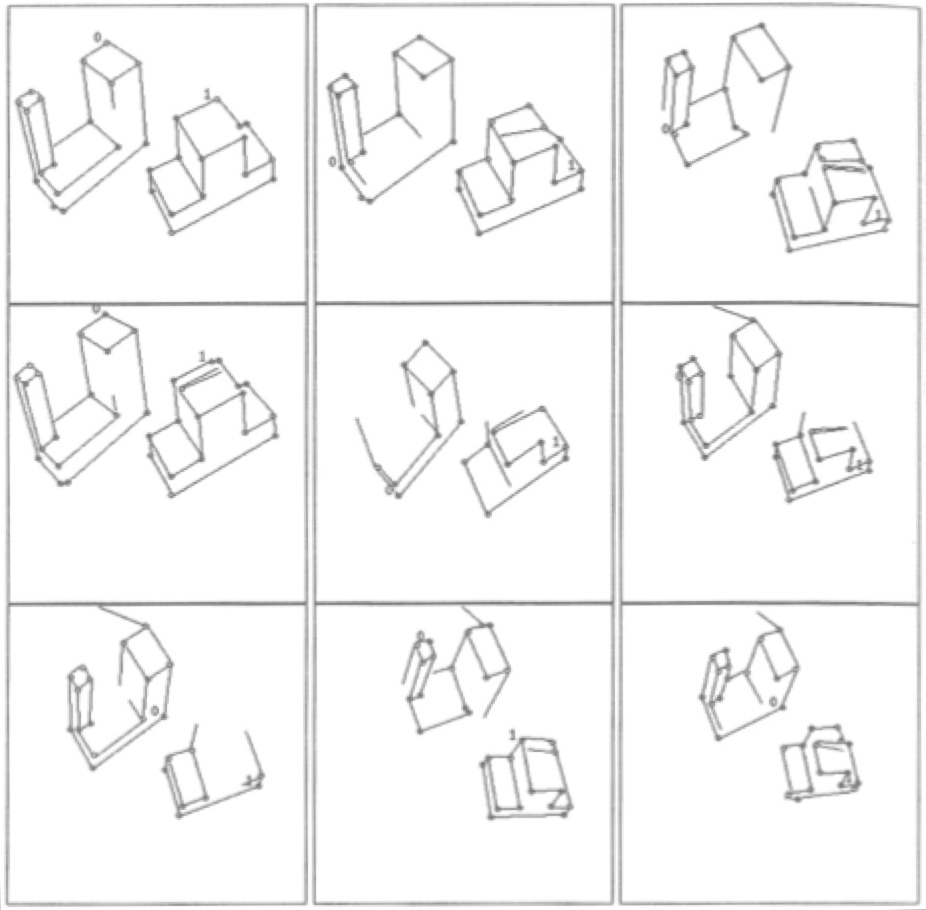}}
\caption{The graphs extracted from the previous images. Notice that the noise
corruption of these graphs varies a lot even if there is only a small change
in camera position and orientation.}
\label{fig:graphs-exp}
\end{figure}

In one such experiment we grabbed 9 images of the same scene and we processed
these images identically (with exactly the same segmentation parameters).
Figure~\ref{fig:images-exp} 
shows these 9 images where the camera position and orientation
varies with respect to the observed scene. Figure~\ref{fig:graphs-exp} 
shows the graphs
extracted from these images. The images are numbered 1 to 9 from left to right
and top to bottom. 

Table~\ref{table:recognition} summarizes the results of recognizing the two
objects based on the two graphs (labelled ``0" and ``1") extracted from the 9
images. The figures in this table correspond to the scores (number of votes)
received by each characteristic view when the image graphs are indexing the
database. For each image graph the table records its highest score, sometimes
the two highest scores. The first object (the graph labelled ``0") has been
correctly recognized 7 times and incorrectly recognized twice (images 5 and
9). Notice the high scores (between 16 and 24) obtained in the case of a
correct recognition in comparison with the less high scores (between 6 and 8)
obtained in the case of an incorrect recognition.

The same phenomena can be observed with the second object (the image graph
labelled ``1") which has been correctly recognized 6 times (the score varies
between 8 and 25) and incorrectly recognized 3 times (the score varies between
4 and 7). An interesting remark is that all 5 incorrect recognitions assigned
{\bf the same} characteristic view to the unknown image graphs, namely the
view labelled ``pieceLd02m" (the bottom-rightmost view on Figure~\ref{fig:vue-caract-edges}). One may notice the poor segmentation associated with this
characteristic view.

The recognition results reported above are barely affected if one increased
the size of the database of characteristic views by adding views of very
different objects. Of course, if the database contains two very similar
objects, the system will fail to discriminate between these two objects.

\begin{table}[h!]
\begin{center}
\resizebox{0.5\textwidth}{!}{
\begin{tabular}{||c|c|c|c|c|c|c|c||}
\hline
image & graph & \multicolumn{6}{c||}{characteristic view} \\
\cline{3-8}
number & number  & Ut01m & Ut02m & M01m & M02m & L05m & Ld02m \\
\hline
1     &  0    & 18$^{\star}$    &  24$^{\star}$   &    &      &      &       \\
\cline{2-8}
      &  1    &       &       &  23$^{\star}$  &      &      &       \\
\hline
2     &  0    & 16$^{\star}$   &  16$^{\star}$   &      &      &      &       \\
\cline{2-8}
      &  1    &       &       &  17$^{\star}$  &      &      &       \\
\hline
3     &  0    &      &  18$^{\star}$   &      &      &      &       \\
\cline{2-8}
      &  1    &       &       &   8$^{\star}$  &      &      &       \\
\hline
4     &  0    & 17$^{\star}$   &  24$^{\star}$   &      &      &      &       \\
\cline{2-8}
      &  1    &       &       &  13$^{\star}$  &      & 10   &       \\
\hline
5     &  0    &      &       &      &  6   &      &  7    \\
\cline{2-8}
      &  1    &       &       &  9$^{\star}$   &      &      &       \\
\hline
6     &  0    & 17$^{\star}$   &  23$^{\star}$   &      &      &      &       \\
\cline{2-8}
      &  1    &       &       &  6   &      &      & 7     \\
\hline
7     &  0    &      &  23$^{\star}$   &      &      &      &       \\
\cline{2-8}
      &  1    &       &       &      &      &      & 4     \\
\hline
8     &  0    &      &  16$^{\star}$   &      &      &      &       \\
\cline{2-8}
      &  1    &       &       &  4   &      &      & 5     \\
\hline
9     &  0    &      &       &      &  6   &      &  8    \\
\cline{2-8}
      &  1    &       &       &  13$^{\star}$  &      &      &       \\
\hline
\end{tabular}}
\end{center}
\caption{This table shows the results of recognition for 9 images of
the same scene. The figures correspond to scores (number of votes) as a result
of the graph indexing process. The scores over-scripted by a $\star$ correspond
to a correct recognition.}
\label{table:recognition}
\end{table}

\section{Discussion}
\label{section:discussion}

Unlike the prevailing paradigm in computer vision that suggests 
image-feature-to-object-feature matching to solve for object recognition,
we described an approach that uses an indexing technique to compare objects
in an image with objects in a database. Our method doesn't rely neither on
precise knowledge about the geometry of the objects nor on reliable
feature-to-feature assignments. Instead we describe both the images and the
models by weighted graphs and we compare these graphs through their polynomial
characterization, namely the second immanantal polynomial of the Laplacian
matrix of a graph. This graph comparison was implemented in two steps
(off-line and on-line):
a database construction step (model graphs are stored in hash tables) and an
indexing step (an image graph indexes in the pre-stored hash tables).

It is worthwhile to notice that, in the past, polynomial characterization of
graphs has been used to represent and identify the topology of molecules
\cite{Kudo73}. At our knowledge, the graph theory literature doesn't describe
any attempt to generalize polynomial characterizations to weighted graphs. It
turns out that, at least for our purposes, this generalization is
straightforward since the Laplacian matrix of a weighted graph has the same
mathematical properties as the Laplacian matrix of a binary graph.

We believe that our indexing scheme based on this algebraic graph
representation  has a promising potential in computer vision and may provide
in the future an interesting paradigm for indexing. 
\section*{Acknowledgements}
This work has been supported by the Esprit programme through the SECOND project (Esprit-BRA No. 6769) and by the ORASIS project (PRC Communication homme/machine).
%\balance
%\bibliography{general,stereo,horaud,tactile,books,rapport01,sensing,gros}

\begin{thebibliography}{10}

\bibitem{Ambler73}
A.P. Ambler, H.G. Barrow, C.M. Brown, R.M. Burstall, and R.J. Popplestone.
\newblock A versatile computer-controlled assembly system.
\newblock In {\em Proc. Third International Joint Conference on Artificial
  Intelligence}, pages 298--307, Stanford University, CA, USA, August 1973.

\bibitem{BallardBrown82}
D.~H. Ballard and C.~M. Brown.
\newblock {\em Computer Vision}.
\newblock Prentice Hall Inc., Englewood Cliffs, New Jersey, 1982.

\bibitem{BollesCain82}
R.~C. Bolles and R.~A. Cain.
\newblock Recognizing and locating partially visible objects, the
  {L}ocal-{F}eature-{F}ocus method.
\newblock {\em International Journal of Robotics Research}, 1(3):57--82, 1982.

\bibitem{BollesHoraud86}
R.~C. Bolles and R.~Horaud.
\newblock 3{DPO}: A three-dimensional part orientation system.
\newblock {\em International Journal of Robotics Research}, 5(3):3--26, Fall
  1986.

\bibitem{FaugerasHebert86}
O.D. Faugeras and M.~Hebert.
\newblock The representation, recognition, and locating of 3-d objects.
\newblock {\em International Journal of Robotics Research}, 5(3):27--52, Fall
  1986.

\bibitem{AyacheFaugeras86}
N.~Ayache and O.~D. Faugeras.
\newblock {HYPER}: {A} new approach for the recognition and positioning of
  two-dimensional objects.
\newblock {\em IEEE Trans. on Pattern Analysis and Machine Intelligence},
  PAMI-8(1):44--54, January 1986.

\bibitem{GrimsonLozanoPerez87}
W.E.L. Grimson and T.~Lozano-Perez.
\newblock Localizing overlapping parts by searching the interpretation tree.
\newblock {\em IEEE Transactions on Pattern Analysis and Machine Intelligence},
  PAMI-9(4):469--482, July 1987.

\bibitem{Goad86}
C.~Goad.
\newblock Fast 3{D} model-based vision.
\newblock In Alex~B. Pentland, editor, {\em From Pixels to Predicates},
  chapter~16, pages 371--391. Ablex Publishing Corporation, Norwood, New
  Jersey, 1986.

\bibitem{FlynnJain91b}
P.~J. Flynn and A.~K. Jain.
\newblock {BONSAI}: 3-{D} object recognition using constrained search.
\newblock {\em IEEE Transactions on Pattern Analysis and Machine Intelligence},
  13(10):1066--1075, October 1991.

\bibitem{KimKak91}
W.~Y. Kim and A.~C. Kak.
\newblock 3-{D} object recognition using bipartite matching embedded in
  discrete relaxation.
\newblock {\em IEEE Transactions on Pattern Analysis and Machine Intelligence},
  13(3):224--251, March 1991.

\bibitem{FlynnJain91}
P.~J. Flynn and A.~K. Jain.
\newblock {CAD}-based computer vision: From cad models to relational graphs.
\newblock {\em IEEE Transactions on Pattern Analysis and Machine Intelligence},
  13(2):114--132, February 1991.

\bibitem{DickinsonPentlandRosenfeld92}
S.~Dickinson, A.~Pentland, and A.~Rosenfeld.
\newblock 3-{D} shape recovery using distributed aspect matching.
\newblock {\em IEEE Transactions on Pattern Analysis and Machine Intelligence},
  14(2):174--198, 1992.

\bibitem{BergevinLevine93}
R.~Bergevin and M.~D. Levine.
\newblock Generic object recognition: Building and matching coarse descriptions
  from line drawings.
\newblock {\em IEEE Transactions on Pattern Analysis and Machine Intelligence},
  15(1):19--36, January 1993.

\bibitem{Umeyama88}
S.~Umeyama.
\newblock An eigendecomposition approach to weighted graph matching problems.
\newblock {\em IEEE Transactions on Pattern Analysis and Machine Intelligence},
  10(5):695--703, May 1988.

\bibitem{Hanajik93}
M.~Hanajik, F.~J. Kylstra, and R.~G. van Vliet.
\newblock An analytical approach to the matching of attributed graphs.
\newblock In {\em Proceedings of the 8th Scandinavian Conference on Image
  Analysis}, pages 419--425, Tromso, August 1993.

\bibitem{AlmohamadDuffuaa93}
H.~A. Almohamad and S.~O. Duffuaa.
\newblock A linear programming approach for the weighted graph matching
  problem.
\newblock {\em IEEE Transactions on Pattern Analysis and Machine Intelligence},
  15(5):522--525, May 1993.

\bibitem{HeraultHoraudVeillonNiez90}
L.~H\'{e}rault, R.~Horaud, F.~Veillon, and J-J. Niez.
\newblock {Symbolic Image Matching by Simulated Annealing}.
\newblock In {\em Proceedings British Machine Vision Conference}, pages
  319--324, Oxford, Great Britain, September 1990.

\bibitem{TrespGindi90}
V.~Tresp and G.~Gindi.
\newblock Invariant object recognition by inexact subgraph matching with
  applications in industrial part recognition.
\newblock In {\em Proc. International Neural Network Conference}, pages 95--98,
  Paris, July 1990.

\bibitem{NevatiaBinford77}
R.~Nevatia and T.~Binford.
\newblock Description and recognition of complex-curved objects.
\newblock {\em Artifitial Intelligence}, 8:77--98, 1977.

\bibitem{Ettinger88}
G.~J. Ettinger.
\newblock {Large Hierarchical Object Recognition Using Libraries of
  Parmeterized Model Sub-Parts}.
\newblock In {\em Proc. Computer Vision and Pattern Recognition}, pages 32--41,
  Ann Arbor, Michigan, USA, June 5-9 1988.

\bibitem{Kalvin86}
A.~Kalvin, E.~Schomberg, J.~T. Schwartz, and M~Sharir.
\newblock Two-dimensional model-based, boundary matchig using footprints.
\newblock {\em The International Journal of Robotics Research}, 5(4):38--54,
  1986.

\bibitem{LamdanWolfson88}
Y.~Lamdan and H.~Wolfson.
\newblock Geometric hashing: A general and efficient model-based recognition
  scheme.
\newblock In {\em Second International Conference on Computer Vision}, pages
  238--249, Tampa, Florida, USA, December 1988.

\bibitem{SteinMedioni92}
F~Stein and G.~Medioni.
\newblock Structural hashing: Efficient 3-d object recognition.
\newblock {\em IEEE Transactions on Pattern Analysis and Machine Intelligence},
  14(2):125--145, February 1992.

\bibitem{Breuel89}
T.~M. Breuel.
\newblock Adaptive model base indexing.
\newblock In {\em Proc. DARPA IU Workshop}, pages 805--814, 1989.

\bibitem{CvetkovicDoobSachs80}
D.~M. Cvetkovic, M.~Doob, and H.~Sachs.
\newblock {\em Spectra of Graphs}.
\newblock Academic Press, New York, 1980.

\bibitem{Turner68}
J.~Turner.
\newblock Generalized matrix functions and the graph isomorphism problem.
\newblock {\em SIAM J. Appl. Math}, 16(3):520--526, May 1968.

\bibitem{MerrisRebmanWatkings81}
R.~Merris, K.~R. Rebman, and W.~Watkings.
\newblock Permanental polynomials of graphs.
\newblock {\em Linear Algebra Applications}, 38:273--288, 1981.

\bibitem{Merris86}
R.~Merris.
\newblock The second immanantal polynomial and the centroid of a graph.
\newblock {\em SIAM Journal Alg. Disc. Meth.}, 7:484--503, 1986.

\bibitem{Constantine90}
G.~M. Constantine.
\newblock Graph complexity and the laplacian matrix in blocked experiments.
\newblock {\em Linear and Multilinear Algebra}, 28:49--56, 1990.

\bibitem{Sedgewick88}
R.~Sedgewick.
\newblock {\em Algorithms}.
\newblock Addison-Wesley Publishing Company, Inc., Massachusetts, 1988.

\bibitem{Gros92c}
P.~Gros and R.~Mohr.
\newblock {Automatic object modelization in computer vision}.
\newblock In H.~Bunke, editor, {\em Proceedings of the workshop ``Advances in
  Structural and Syntactic Pattern Recognition'', Bern, Switzerland}, volume~5
  of {\em Series on Machine Perception and Artificial Intelligence}, pages
  385--400. World Scientific, August 1992.

\bibitem{Gros93d}
P.~Gros.
\newblock {Matching and clustering: two steps towards automatic model
  generation in computer vision}.
\newblock In {\em Proceedings of the AAAI Fall Symposium Series: Machine
  Learning in Computer Vision: What, Why, and How?, Raleigh, North Carolina,
  USA}, pages 40--44, October 1993.

\bibitem{HeraultNiez89}
L.~H\'erault and J.J. Niez.
\newblock {Neural Networks and Graph K-Partitioning}.
\newblock {\em Complex Systems}, 3(6):531--576, December 1989.

\bibitem{Kanade81}
T.~Kanade.
\newblock Recovery of the 3{D} shape of an object from a single view.
\newblock {\em Artificial Intelligence}, 17(1-3):409--460, August 1981.

\bibitem{HoraudVeillonSkordas90}
R.~Horaud, F.~Veillon, and Th. Skordas.
\newblock Finding geometric and relational structures in an image.
\newblock In O.~Faugeras, editor, {\em Computer Vision -- ECCV 90, Proceedings
  First European Conference on Computer Vision, Antibes, France}, pages
  374--384. Springer Verlag, April 1990.

\bibitem{Kudo73}
Y.~Kudo, T.~Yamasaki, and S.~Sasaki.
\newblock The characteristic polynomial uniquely represents the topology of a
  molecule.
\newblock {\em Journal of Chemical Documentation}, 13(4):225--227, 1973.

\end{thebibliography}

\balance
\end{document}